# An explanation method for Siamese neural networks

Lev V. Utkin, Maxim S. Kovalev, and Ernest M. Kasimov

Peter the Great St.Petersburg Polytechnic University (SPbPU), Saint-Petersburg, Russia
`lev.utkin@gmail.com`

**Abstract.** A new method for explaining the Siamese neural network is proposed. It uses the following main ideas. First, the explained feature vector is compared with the prototype of the corresponding class computed at the embedding level (the Siamese neural network output). The important features at this level are determined as features which are close to the same features of the prototype. Second, an autoencoder is trained in a special way in order to take into account the embedding level of the Siamese network, and its decoder part is used for reconstructing input data with the corresponding changes. Numerical experiments with the well-known dataset MNIST illustrate the propose method.

## 1  Introduction

Deep models play an important role in making prediction for many applications. However, a lot of machine learning techniques are not explainable, they are black boxes and do not explain their predictions. This may be a problem for applying the models to various field, for example, to medicine. Therefore, a lot of explanation models have been developed, which can be viewed as special meta-models for explaining the deep model predictions [1,2].

A lot of machine learning models are regarded as black boxes, i.e., it is assumed that we do not know any details of the black-box model, for example, its structure, parameters, etc., but its input and the corresponding output are known and can be used for training the explanation model. A well-known method is the Local Interpretable Model-agnostic Explanations (LIME) [3]. According to the LIME, the explanation may be derived locally from randomly generated synthetic neighbor examples. There are also modifications of the LIME, for example, ALIME [4], NormLIME [5], DLIME [6]. Another well-known method is the SHAP [7] and its modifications [8,9]. It should be noted that there are also alternative methods, for example, influence functions [10], a multiple hypothesis testing framework [11], counterfactual explanations [12].

Some explanation methods use a prototype technique which selects representative instances (prototypes) from the training data, for instance, from examples belonging to the same class [13,14].

Another interesting idea realized in many explanation methods is the perturbation technique[15,16]. These methods assume that contribution of a feature can be determined by measuring how prediction score changes when the feature is altered [7]. At the same time, the perturbation technique may be computationally hard when perturbed inputs have a lot of features, for example, pictures.

We have to point out a number of interesting survey papers devoted to explainable methods [17,18,19], which cover many questions related to the methods.

We consider an approach which is agnostic to the black-box model. This means that we do not know or do not use any details of the black-box model. Only its input and the corresponding output are used for training the explanation model.

To the best of our knowledge, there are no appropriate algorithms for explaining the Siamese neural network (SNN). Therefore, we propose a method to explain the SNN [20,21] as the black-box model. The SNN consists of two identical neural subnets sharing the same set of weight. The SNN aims to compare a pair of feature vectors in terms of their semantic similarity or dissimilarity. It realizes a non-linear embedding of data with the objective to bring together similar examples and to move apart dissimilar examples. SNNs have been applied to various problems, including image recognition and verification, visual tracking, novelty and anomaly detection, one-shot and few-shot learning [21-27].

Problems for explaining the SNN stem from two main reasons. First, the input examples for the SNN are semantically similar or dissimilar, and direct distances between them may not have a sense. Second, there is no an inverse map between the embeddings and the corresponding input examples, i.e., we do not know subsets of features in the input vector corresponding to some features of the embedded vector.

We try to solve these problems by applying the following ideas. First, we find prototypes of all classes at the embedding level and select features having the smallest Euclidean distance between the embedding of the explained example and the prototype. Second, we train an autoencoder with a special loss function which takes into account embeddings obtained by means of the SNN. The decoder part of the autoencoder is used to reconstruct the introduced perturbations to observe features of the reconstructed example with largest changes. These features are nothing else, but the explanation of interest.

The paper is organized as follows. A description of the SNN and its peculiarities are given in Section 2. Two important concepts of explanation methods, the perturbation technique and prototypes, are considered in the same section. The proposed explanation method, which is the aim of the paper, is provided in Section 3. Numerical experiments illustrating the proposed method on the basis of the well-known MNIST dataset are studied in Section 4. Concluding remarks are provided in Section 5.

## 2    Siamese neural networks, perturbations and prototypes

Let $\{(\mathbf{x}_i, y_i), i=1,...,n\}$ be a dataset consisting of $n$ feature vectors $\mathbf{x}_i \in \mathbf{R}^m$ of size $m$ with labels $y_i \in \{1,2,...,C\}$. Let us construct a new training set $S = \{(\mathbf{x}_i, \mathbf{x}_j, z_{ij}), (i,j) \in K\}$ consisting of pairs of examples $\mathbf{x}_i$ and $\mathbf{x}_j$ with binary labels $z_{ij} \in \{0,1\}$ assigned to them. If both feature vectors $\mathbf{x}_i$ and $\mathbf{x}_j$ are semantically similar, i.e., they belong to the same class, then $z_{ij}$ is 0. If the vectors are semantically dissimilar, i.e., they correspond to different classes, then $z_{ij}$ is 1. So, the training set $S$ can be divided into two subsets: a similar or positive set with $z_{ij}=0$ and a dissimilar or negative set with $z_{ij}=1$.

Suppose new feature representations of the input examples $\mathbf{x}_i$ and $\mathbf{x}_j$ as the SNN outputs are $\mathbf{h}_i \in \mathbf{R}^D$ and $\mathbf{h}_j \in \mathbf{R}^D$, respectively. The SNN realizes a map $f$ such that $\mathbf{h}_i = f(\mathbf{x}_i)$, which tries to make the Euclidean distance $d(\mathbf{h}_i, \mathbf{h}_j)$ as small (large) as possible for the similar (dissimilar) pair of objects. A standard architecture of the SNN is shown in Fig. 1.

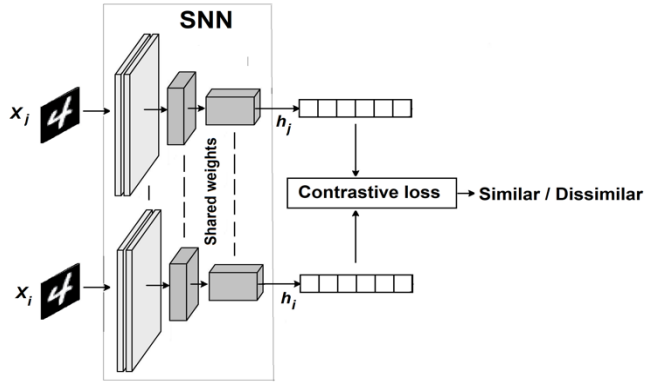

**Fig. 1.** An architecture of the SNN.

It should be noted that there are many specific loss functions for training the SNN [28]. We use a function called the contrastive loss function. It is defined as

$$l(\mathbf{x}_i, \mathbf{x}_j, z_{ij}) = (1 - z_{ij})\|\mathbf{h}_i - \mathbf{h}_j\|_2^2 + z_{ij} \max(0, \tau - \|\mathbf{h}_i - \mathbf{h}_j\|_2^2).$$

Here $\tau$ is a predefined threshold. Hence, the total error function for minimizing is defined as

$$L(W) = \sum_{i,j} l(\mathbf{x}_i, \mathbf{x}_j, z_{ij}) + \mu R(W).$$

Here $R(W)$ is a regularization term added to improve generalization of the neural network; $W$ is the matrix of the neural subnet parameters; $\mu$ is a hyper-parameter which controls the strength of the regularization. The above problem is usually solved by using a gradient descent scheme.

Let us consider definitions of perturbations and prototypes, which will be used in the proposed explanation method. It has been mentioned that some explanation methods are based on applying perturbation schemes which explicitly test the explained model's response to local perturbations. The intuition behind the technique is that the more a model's response depends on a feature, the more predictions change with the corresponding feature changes. The perturbation scheme $\rho$ perturbs all features of $\mathbf{x}$ into $\hat{\mathbf{x}}$ as follows: $\rho(\mathbf{x}, \delta) = \hat{\mathbf{x}} = \mathbf{x} + \delta$. Here $\delta$ is the perturbation vector. In many cases, finding the optimal perturbing scheme for all instances is intractable due to a possible large dimensionality of input examples. Therefore, various techniques are available to simplify this procedure.

Following the work of Snell et al. [29], a prototype is a data example that is representative of a subset of data, for example, a set of examples from a class. If we have $D$-dimensional representation of every $\mathbf{x}_i$ through an embedding function $f : \mathbf{R}^m \to \mathbf{R}^D$, then the prototype $\mathbf{c}_k \in \mathbf{R}^D$ of class $k$ is defined as [29]:

$$\mathbf{c}_k = \frac{1}{n_k} \sum_{i \,:\, y_i = k} f(\mathbf{x}_i) = \sum_{i \,:\, y_i = k} \mathbf{h}_i.$$

## 3 The proposed method for the SNN explanation

The proposed method for explaining the SNN can be represented by means of an algorithm consisting of two parts. The first part aims to train the additional autoencoder with a special loss function. This autoencoder can be called as an explainable autoencoder. The goal of the second part is to perturb the embedding vector at the SNN output and to use the decoder of the trained autoencoder in order to reconstruct the perturbed embeddings and to observe the features which are changed.

Suppose we have a trained SNN as a black box. For every input vector $\mathbf{x}_i$, we have the corresponding embedding vectors $\mathbf{h}_i$ such that $\mathbf{h}_i = f(\mathbf{x}_i)$. The main idea to incorporate the additional autoencoder is the following.

Suppose we have an embedding vector $\mathbf{h}$ with a set of important features. However, these features do not explain why the considered explained example belongs to a class, say to class $k$. In order to answer this question, it is necessary to find an inverse mapping from $\mathbf{h}$ to $\mathbf{x}$, i.e., the vector $\mathbf{x}$ has to be reconstructed from $\mathbf{h}$. The reconstruction can be carried out by means of a neural network with input values $\mathbf{h}_i$ and output values $\mathbf{x}_i$. Our numerical experiments have demonstrated that it is difficult to train a reconstruction neural network especially when the dimension of vectors $\mathbf{x}$ is very large and the number of training examples is small due to possible overfitting of the network. It is simpler to train an autoencoder and then to use its trained decoder part for reconstruction. In order to exploit the decoder for reconstruction of vector $\mathbf{h}$, the autoencoder has to be trained in a special way. First of all, the length $D$ of its code (the hidden representation) has to coincide with the length of vector $\mathbf{h}$. The loss function should take into account the proximity of vectors $\mathbf{h}_i$ and the corresponding vectors of the autoencoder hidden representation.

Suppose that the input examples for the proposed incorporated autoencoder are vectors $\mathbf{x}_i$, then we expect to get reconstructed vectors $\tilde{\mathbf{x}}_i$ as its outputs. The corresponding loss function $L_{\text{recon\_a}}$ for training the autoencoder is defined as follows:

$$L_{\text{recon\_a}}(W) = \sum_{i=1}^{n} \|\mathbf{x}_i - \tilde{\mathbf{x}}_i\|_2^2.$$

We do not write a regularization term because it will be used later. However, we cannot apply the autoencoder in its standard form because we need to have the vectors $\mathbf{h}_i^*$ in the hidden layer coinciding with the vectors $\mathbf{h}_i$ obtained by means of the SNN. Therefore, we propose to change the loss function for training the autoencoder by adding the loss function $L_{\text{close}}$ in the following way:

$$L_{\text{autoen}}(W) = \gamma L_{\text{recon\_a}}(W) + \mu L_{\text{close}}(W) + \lambda R(W)$$

$$= \gamma \sum_{i=1}^{n} \|\mathbf{x}_i - \tilde{\mathbf{x}}_i\|_2^2 + \mu \sum_{i=1}^{n} \|\mathbf{h}_i - \tilde{\mathbf{h}}_i\|_2^2 + \lambda R(W).$$

Here $R(W)$ is a regularization term, $\lambda$ is a hyper-parameter which controls the strength of the regularization; $W$ is the set of the neural network weights; $\gamma$ and $\mu$ are weights that control the interaction of the loss terms; $\tilde{\mathbf{h}}_i$ are the autoencoder hidden representation vectors.

We can now use the decoder part for reconstruction of the perturbed embeddings that is for implementing the second part of the algorithm. This trick allows us to significantly simplify the training process and to get acceptable vector reconstructions. It should be noted that an architecture of the encoder differs from the architecture of a subnetwork of the SNN because we consider the SNN as a black box whose architecture is unknown. A scheme of the explanation algorithm first part is shown in Fig. 2. It can be seen from the training scheme that the autoencoder is trained by using embedding vectors from one of the SNN subnetworks.

Now we consider the second part of the explanation algorithm under condition that there is available the trained decoder for reconstruction. A schematic representation of the part is shown in Fig. 3. It is based on using prototypes and perturbations. By having embedding vectors $\mathbf{h}_i$ for all training examples $\mathbf{x}_i$, we can compute prototypes $\mathbf{c}_k \in \mathbf{R}^D$ for every class $k$ as it is shown in Section 2 based on embedded vectors $\mathbf{h}_i$ such that $y_i = k$. Without loss of generality, we suppose that an explained example $\mathbf{x}$ belongs to class $k$. It is obvious in this case, that the explained example and the prototype are semantically similar (of course if the SNN is correctly classified the example). This implies that the Euclidean distance between the embedded vector $\mathbf{h}$ of the explained example and the prototype $d(\mathbf{h},\mathbf{c}_k)$ should be smaller than the Euclidean distance between the embedded vector of the explained example and the prototypes of other classes $d(\mathbf{h},\mathbf{c}_i)$, $i \neq k$.

So, we have vectors $\mathbf{h}$ and $\mathbf{c}_k$ consisting of $D$ features. It is obvious that features in $\mathbf{h}$, which are close to the corresponding features in $\mathbf{c}_k$, can be viewed as important features. These important features define the fact that the explained example belongs to class $k$. Therefore, they should be selected. Let us introduce the number of important features $s < D$ such that the index set $J \subseteq \{1,...,D\}$ consists of $s$ indices corresponding to smallest distances between $h_i$ and $c_i^{(k)}$, $i = 1,..,D$. Denote ordered $s$ features of $\mathbf{h}$ as $h_{(i)}$, $i = 1,...,s$. Then by perturbing the embedded vector $\mathbf{h}$ (features $h_{(1)},..., h_{(s)}$), we can construct a training set which consists vectors $\mathbf{h}+\delta_i$. Here $\delta_i$ is the perturbation vector such that indices of its non-zero elements are from the index set $J$, other elements are equal to zero.

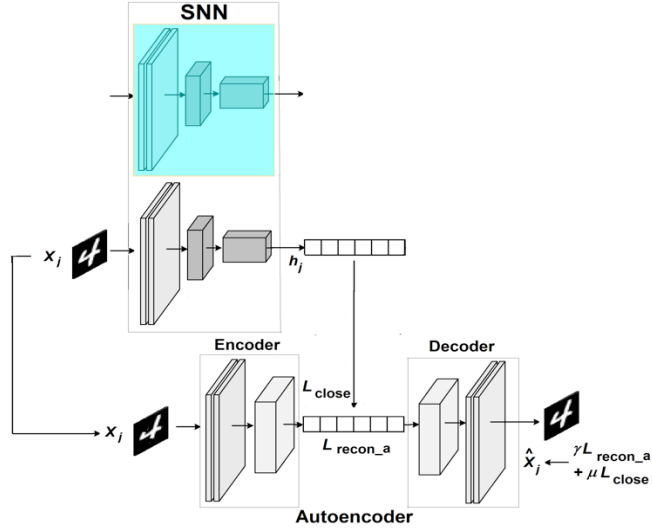

**Fig. 2.** The autoencoder training part of the explanation algorithm.

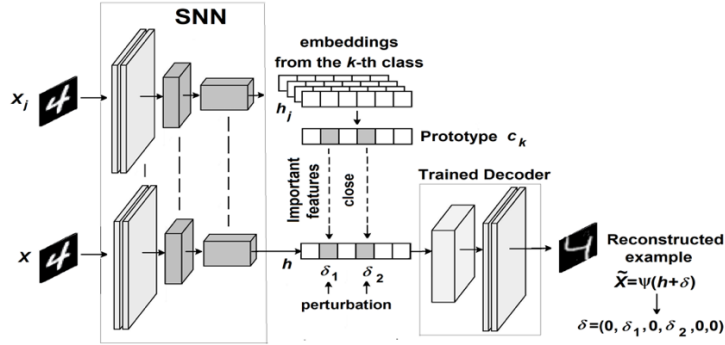

**Fig. 3.** The second part of the explanation algorithm.

In order to study how the important features of the hidden representation impact on the original vector $\mathbf{x}$, we use the trained decoder to reconstruct vectors from $\mathbf{h}+\boldsymbol{\delta}_i$ and to investigate how features of the reconstructed vector $\tilde{\mathbf{x}} \in \mathbf{R}^m$ are changed. In sum, we have the embedding vector $\mathbf{h}$, the reconstruction $\tilde{\mathbf{x}}$, the index set $J$ of important features of $\mathbf{h}$. The trained decoder implements a function $\psi : \mathbf{R}^D \to \mathbf{R}^m$, i.e., $\tilde{\mathbf{x}} = \psi(\mathbf{h})$. In order to determine the important features of $\tilde{\mathbf{x}}$, we perturb the important features of $\mathbf{h}$ and to observe which features of $\tilde{\mathbf{x}}$ have the largest changes. We use the random perturbation of important features of $\mathbf{h}$ when every feature is added with a positive random value. Moreover, the pre-trained decoder is again trained by using only the SNN output embeddings.

Let $q$ be the number of important features $\tilde{\mathbf{x}}$. By generating the random vector $\boldsymbol{\delta}$ many times, say $N$ times, and observing changes of $\tilde{\mathbf{x}}$, we can compute mean realize changes of all features. Then features, having $q$ largest changes of the reconstructed example, explain the considered example.

Let us return to the scheme in Fig. 3. It can be seen from the scheme that one sub-network in the SNN is conditionally used for getting the vector $\mathbf{h}$. Another subnetwork provides a set of embedding vectors in order to compute the prototype $\mathbf{c}_k$. The important features (two features) in $\mathbf{h}$ are shown by dashed cells. They are close to the same features in the prototype. The reconstruction network (decoder) is shown on the right side of the picture. The perturbed vector is fed to the decoder in order to get the reconstructed vector which depends on perturbations.

Perturbation vectors are randomly generated with respect to the normal distribution with zero expectations and small variances of $s$ features in accordance with the index set $J$. We take only positive perturbations because changes of features should be closer to the prototype.

## 4   Numerical experiments

The proposed explanation method is studied by applying the MNIST dataset which is a commonly used large dataset of 28x28 pixel handwritten digit images [30]. It has a training set of 60,000 examples, and a test set of 10,000 examples. The digits are size-normalized and centered in a fixed-size image. The dataset is available at http://yann.lecun.com/exdb/mnist/.

The length of the hidden representation layer is 10, i.e., the vector $\mathbf{h}$ consists of 10 features. The autoencoder implementation uses ReLu as activation function for all layers except for the last layer where a sigmoid activation function is used.

Perturbations are sampled from the normal distribution with zero mean and standard deviation $0.1 \min\{|h_1 - c_{k1}|, \ldots, |h_D - c_{kD}|\}$, where $h_i$ is the $i$-th component of vector $\mathbf{h}$, $c_i^k$ is the $i$-th component of the prototype $\mathbf{c}_k$. The number of perturbations is 5000.

An architecture of the autoencoder is given in Table 1. The architecture contains an encoder (the first and second columns) and an equivalent decoder (the third and fourth columns). The encoder comprises convolution layers (Conv), max pooling operations (Pooling), flatten layers (Flatten) which flatten a matrix input to a simple vector, dense layers (Dense) which are a fully connected layer. The decoder block has additionally deconvolution layers (UpSampling), reshape layers (Reshape) which change the dimensions of its input without changing its data.

We show below quadruples of pictures such that the first picture in every quadruples is an original image of a digit, the second picture is the reconstructed digit, the third picture is the original image and the corresponding mask of explanation features, the fourth picture is the explanation feature in the form of the mask. The explanation features can be regarded as correct if they clearly show difference of the considered digit belonging to the classified class from digits belonging to other classes.

**Table 1.** The autoencoder architecture.

| Encoder | | Decoder | |
| --- | --- | --- | --- |
| Layer | Output | Layer | Output |
| Input | 28x28x1 | Input | 20 |
| Conv1 | 28x28x16 | Dense1 | 40 |
| Pooling1 | 14x14x16 | Dense2 | 128 |
| Conv2 | 14x14x8 | Reshape | 4x4x8 |
| Pooling2 | 7x7x8 | UpSampling1 | 8x8x8 |
| Conv3 | 7x7x8 | Conv1 | 7x7x8 |
| Pooling3 | 4x4x8 | UpSampling2 | 14x14x8 |
| Flatten | 128 | Conv2 | 14x14x16 |
| Dense1 | 40 | UpSampling3 | 28x28x16 |
| Dense2 | 20 | Conv3 | 28x28x1 |

Four examples of the correct explanation of digits from MNIST are shown in Figs. 4-7. It can be seen from the pictures that all original digits are perfectly reconstructed by the trained decoder. However, the quality of explanation depends on the reconstructed images. Fig. 8 illustrates an example of the incorrect explanation when the reconstructed image significantly differs from the original image. This implies that the autoencoder is not perfectly trained or its architecture does not allow us to efficiently reconstruct all images from the testing set. Another interesting case is when the digits are incorrectly classified by the SNN. This case is demonstrated in Fig. 9, where the original digit 4 is classified by the SNN as the digit 9. As a result, the explainer selects features which actually indicate the digit 9 instead of 4. In fact, this case shows that the proposed method correctly explains, but the explanation depends on the correctness of the black-box model classification.

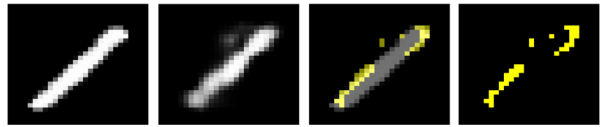

**Fig. 4.** Explanation of the digit 1.

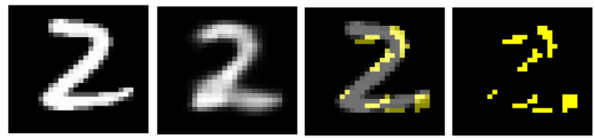

**Fig. 5.** Explanation of the digit 2.

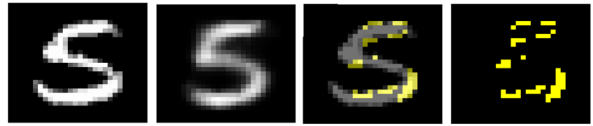

**Fig. 6.** Explanation of the digit 5.

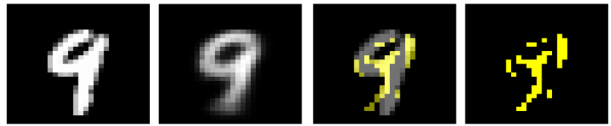

**Fig. 7.** Explanation of the digit 9.

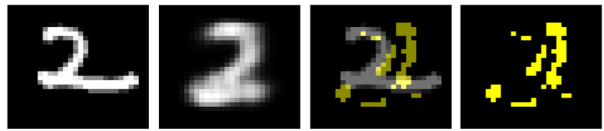

**Fig. 8.** The incorrect explanation of the digit 2.

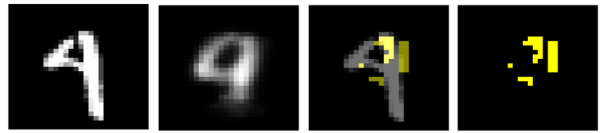

**Fig. 9.** The digit 4 is incorrectly classified to 9.

## 5    Conclusion

A new method for explaining the SNN has been presented in the paper. The main ideas underlying the method are comparison of the explained example with a prototype at the embedding level and reconstruction of the embedding feature vectors by means of a separately trained special autoencoder in order to analyze the impact of the embedding vector perturbations on the reconstructed features. The proposed method can be applied to various problems which use SNNs.

It is important to note that the SNN can be regarded as a part of a general distance metric learning approach. Therefore, applications of the proposed explanation method can be extended on various models of the distance metric learning. One of the interesting directions for the extension is the novelty and anomaly detection because this problem has a huge amount of applications.

A bottleneck of the proposed model is the autoencoder which has to be trained by using the dataset. The problem arises when the analyzed dataset is rather small. Ways for solving the problem can be regarded as a direction for further research. Another problem is that the method is based on the random perturbations. At the same time, there are a lot of interesting works, for example, [31] or [32], where perturbations are determined in an optimal way by solving the corresponding optimization problems. The use of this approach to modifying the proposed method is another direction for further research.

## 6 Acknowledgement

This work is supported by the Russian Science Foundation under grant 18-11-00078.